\documentclass[conference]{IEEEtran}
\IEEEoverridecommandlockouts
\usepackage{cite}
\usepackage{booktabs}
\usepackage{amsmath,amssymb,amsfonts}
\usepackage{algorithmic}
\usepackage{graphicx}
\usepackage{textcomp}
\usepackage{xcolor}
\usepackage{multirow}
\newcommand{\bre}{\begin{bf}\begin{color}{blue}}
\newcommand{\ere}{\end{color} \end{bf}}

\def\BibTeX{{\rm B\kern-.05em{\sc i\kern-.025em b}\kern-.08em
    T\kern-.1667em\lower.7ex\hbox{E}\kern-.125emX}}
\begin{document}

\title{Schizophrenia Detection from EEG Signals: A Transformer Framework with Spectrogram Representation
}

\author{
\IEEEauthorblockN{1\textsuperscript{st} Abtin Shafiei}
\IEEEauthorblockA{\textit{Department of Computer Sciences} \\ \textit{and Information Technology} \\
\textit{Institute for Advanced Studies} \\ \textit{in Basic Sciences}\\
Zanjan, Iran \\
abtinshf@iasbs.ac.ir}
\and
\IEEEauthorblockN{2\textsuperscript{nd} Mohsen Hooshmand}
\IEEEauthorblockA{\textit{Department of Computer Sciences} \\ \textit{and Information Technology} \\
\textit{Institute for Advanced Studies} \\ \textit{in Basic Sciences}\\
Zanjan, Iran \\
mohsen.hooshmand@iasbs.ac.ir}
\and
\IEEEauthorblockN{3\textsuperscript{rd} Majid Ramezani}
\IEEEauthorblockA{\textit{Department of Computer Sciences} \\ \textit{and Information Technology} \\
\textit{Institute for Advanced Studies} \\ \textit{in Basic Sciences}\\
Zanjan, Iran \\
ramezani@iasbs.ac.ir}
\and

}

\maketitle

\begin{abstract}
% Schizophrenia is a serious psychiatric disorder that affects millions of people worldwide, and its diagnosis remains primarily dependent on clinical assessment. Electroencephalography (EEG) provides a non-invasive approach to investigate brain activity and has shown potential to support automated Schizophrenia detection. However, existing EEG-based classification studies often suffer from limitations including small datasets, inconsistent preprocessing strategies, and evaluation protocols that may not adequately prevent subject-related data leakage. In this study, we propose an EEG-based Schizophrenia classification framework that transforms preprocessed EEG recordings into time-frequency representations using the Short-Time Fourier Transform. The generated spectrogram images are classified using both conventional Machine Learning algorithms, including Support Vector Machines, Random Forests, and XGBoost, and Deep Learning models, including convolutional architectures and CNN-Transformer hybrids. To ensure reliable evaluation, all data partitions are performed at the subject level. Experimental results demonstrate that the proposed approach achieves competitive classification performance, with the CNN-Transformer (CT-SZ) model achieving an AUC-ROC of 88.41\% and the CNN + Squeeze and Excitation + Transformer (CST-SZ) achieving an AUC-ROC of 92.88\% on the independent test set.
Schizophrenia is a serious psychiatric disorder that affects millions of people worldwide, and its diagnosis remains primarily dependent on clinical assessment. Electroencephalography (EEG) provides a non-invasive approach to investigate brain activity and has shown potential to support automated Schizophrenia detection. However, existing EEG-based classification studies often suffer from limitations including small datasets, inconsistent preprocessing strategies, and evaluation protocols that may not adequately prevent subject-related data leakage. In this study, we propose an EEG-based Schizophrenia classification framework that transforms preprocessed EEG recordings into time-frequency representations using the Short-Time Fourier Transform. The generated spectrogram images are classified using both conventional Machine Learning algorithms, including Support Vector Machines, Random Forests, and XGBoost, and Deep Learning models, including convolutional architectures and CNN-Transformer hybrids. To ensure reliable evaluation, all data partitions are performed at the subject level, and image-level predictions are aggregated into subject-level decisions. On the independent test set of 18 subjects, the CNN-Transformer (CT-SZ) model achieves a subject-level AUC-ROC of 95.00\%, while the CNN + Squeeze-and-Excitation + Transformer (CST-SZ) model achieves 92.50\%.
% ; at the image level, CT-SZ and CST-SZ reach 91.51\% and 92.88\%, respectively.

\end{abstract}


\section{Introduction}
Automated Schizophrenia diagnosis using EEG signals has received increasing attention in recent years thanks to advances in Machine Learning and Deep Learning fields. A wide variety of approaches have been proposed, ranging from traditional classifiers trained on handcrafted features to deep neural networks capable of learning discriminative representations directly from EEG-derived inputs. Despite these advances, the literature remains highly heterogeneous with respect to datasets, preprocessing techniques, feature engineering methods, and validation strategies, emphasizing the need for standardized evaluation and reproducible methodologies~\cite{rahul24}.
Schizophrenia is a severe and chronic psychiatric disorder characterized by disturbances in perception, cognition, emotional regulation, and behavior. It represents a significant global public health challenge due to its persistent impact on individuals, families, and healthcare systems. A recent systematic review and meta-analysis of global epidemiological studies demonstrated that Schizophrenia affects populations worldwide, with a consistent lifetime prevalence across diverse geographic regions, highlighting its substantial global burden \cite{zhang26}.

A wide range of Machine Learning and Deep Learning approaches have been proposed for the accurate and efficient detection of Schizophrenia~\cite{bhadra24, aich25}. The majority of these methods rely on EEG signals because of their non-invasive nature and ability to capture neural activity with high temporal resolution. Rather than using raw EEG signals directly, many studies transform them into alternative representations that better characterize their underlying discriminative features, thereby improving classification performance and downstream tasks. Common transformations include frequency-domain representations~\cite{jahmunah19} and time-frequency representations such as scalograms~\cite{khare21}.

Although these approaches have demonstrated promising performance in Schizophrenia detection, developing reliable and generalizable models remains a significant challenge. In this work, we propose a family of CNN-Transformer-based architectures for automated Schizophrenia detection. The proposed methods are evaluated on the ASZED-153 dataset~\cite{mosaku25}, which comprises EEG recordings from a relatively large cohort of subjects. Our framework represents raw EEG signals as time–frequency spectrograms, providing a two-dimensional representation from which discriminative temporal and spectral patterns can be learned for schizophrenia detection.

% Unlike previous studies, our framework employs spectrogram representations of raw EEG signals, enabling the extraction of discriminative time-frequency features that effectively capture the underlying characteristics of the EEG data.

Furthermore, reliable evaluation in medical ML applications requires rigorous data partitioning protocols to prevent information leakage and ensure unbiased performance estimation. Therefore, all training, validation, and testing splits in this study were performed at the subject level, ensuring that EEG recordings from the same individual were exclusively assigned to a single subset.

The main contributions of this work are summarized as follows.
\begin{itemize}
\item Generating spectrogram representations from raw EEG signals for Schizophrenia detection.
\item Proposing a family of CNN-Transformer-based architectures for automated Schizophrenia detection.
\item Evaluating the proposed methods alongside conventional Machine Learning techniques and state-of-the-art approaches from the literature on the ASZED-153 dataset.
\item Providing a fair comparison of the evaluated approaches using a subject-level data partitioning protocol.
\end{itemize}

The remainder of this paper is organized as follows. Section~\ref{sec:rel} reviews the related work on Schizophrenia detection. Section~\ref{sec:prop} describes the ASZED-153 dataset, the proposed methodology, and the experimental pipeline. Section~\ref{sec:res} presents the experimental results and compares the proposed methods with conventional Machine Learning models and existing approaches from the literature. Finally, Section~\ref{sec:conc} concludes the paper and outlines potential directions for future research.

\section{Related work}
\label{sec:rel}
Khare et al. \cite{khare21} developed an automated Schizophrenia detection system by utilizing Time-Frequency Representation (TFR) techniques in conjunction with Convolutional Neural Networks (CNNs) on a dataset containing sensory task recordings of 81 subjects (49 case subjects and 32 control subjects). Their methodology applied band-pass filtered EEG signals into 2-D image representations using the Short-Time Fourier Transform, Continuous Wavelet Transform (CWT), and Smoothed Pseudo-Wigner–Ville Distribution (SPWVD). Then, the resulting three images are stacked together and utilized as inputs for a four-layered CNN architecture. The framework specifically analyzes three distinct press-button task conditions to identify the optimal auditory and motor task-based input, with training and testing performed using 10-fold cross-validation.
Bhadra and Kumar~\cite{bhadra24} proposed an automated EEG-Based Schizophrenia detection pipeline that used Discrete Wavelet Transform (DWT) with both Machine and Deep Learning approaches. They first preprocessed resting-state EEG recordings using artifact removal, applying 2-45 HZ band-pass filtering, and segmenting the signals into thirty-second windows. The signals are then decomposed by a five-level DWT with the Daubechies-4 (db4) wavelet, capturing both the spectral and temporal characteristics. The Principal Component Analysis (PCA) is then applied to reduce the high-dimensionality of the extracted wavelet coefficients.
% This study highlights the effectiveness of wavelet-based feature extraction, and demonstrates that Deep Learning models can leverage DWT-derived features to improve classification, although the authors acknowledge that the relatively small dataset limits the generalizability of the reported results.%
Aich et al.~\cite{aich25} proposed a three-stage hybrid framework for the automated detection of Schizophrenia from EEG signals by transforming the diagnostic task into an image classification problem. The methodology begins by encoding raw EEG time-series data into 2-D scalogram images using the Continuous Wavelet Transform (CWT) with a Morlet wavelet. These images serve as inputs for feature extraction through transfer learning, utilizing pre-trained EfficientNetB3 and DenseNet169 architectures. To address high dimensionality and potential redundancy, the authors introduced an Average Subtraction-based Optimization (ASBO) algorithm. This wrapper-based feature selection method iteratively navigates the search space to identify an optimal feature subset, employing a K-Nearest Neighbors (KNN) classifier as the evaluation mechanism to refine the extracted deep features without requiring specific parameter tuning.
%===============================================================================================================%
Jangde and Verma~\cite{jangde26} proposed an attention-integrated one-dimensional CNN for Schizophrenia detection from resting-state EEG signals. Their model combines one-dimensional convolutional layers with a Convolutional Block Attention Module (CBAM)~\cite{woo18} to emphasize informative EEG channels and temporal regions. Preprocessing includes 1-40 Hz band-pass filtering, Independent Component Analysis (ICA) based artifact removal, 50 Hz notch filtering, Z-score normalization, and overlapping five-second segmentation. The model was evaluated on the IBIB PAN (19 channels, 250 Hz, 28 subjects) and Moscow (16 channels, 128 Hz, 84 subjects) datasets. Results showed that the attention mechanism improved discriminative feature extraction and enhanced interpretability by identifying the most relevant brain regions and temporal segments. 
%===============================================================================================================%
Hossain and Tawhid~\cite{hossain26} proposed an explainable EEG-based Schizophrenia detection framework that combines time-frequency analysis with Deep Learning. Their methodology preprocesses EEG signals through band-pass filtering, resampling, and segmentation before converting them into STFT-based mel-spectrogram images. The EEG channels are further grouped into frontal, temporal, parietal, central, and occipital lobes to enable region-specific analysis alongside full-brain classification. A CNN is trained on the generated spectrograms using 10-fold cross-validation, while Grad-CAM, SHAP, and LIME are employed to interpret the model's predictions and highlight the contribution of different brain regions. 
% This framework integrates lobe-specific analysis with explainable AI to improve the interpretability of automated Schizophrenia diagnosis. %===============================================================================================================%
% \color{red}
Recent hybrid approaches have also explored EEG-based schizophrenia detection using adaptive signal decomposition and machine learning. Sravanthi and Sharma ~\cite{sravanthi26} proposed a framework combining Variational Mode Decomposition (VMD), multi-domain entropy and fractal features, and optimized machine-learning classifiers, achieving accuracies of 96.7\% and 99.0\% on the MHRC and RepOD datasets, respectively. Importantly, the study employed a subject-wise Leave-One-Out Cross-Validation protocol specifically to prevent data leakage, enabling a more direct assessment of subject-independent generalization. In contrast, we evaluate all models, including a reimplemented DWT-based baseline~\cite{bhadra24}, under a unified subject-level protocol.
% \color{black}

\section{Proposed method}
\label{sec:prop}
\begin{figure*}[t]
    \centering
    \includegraphics[width=0.8\textwidth]{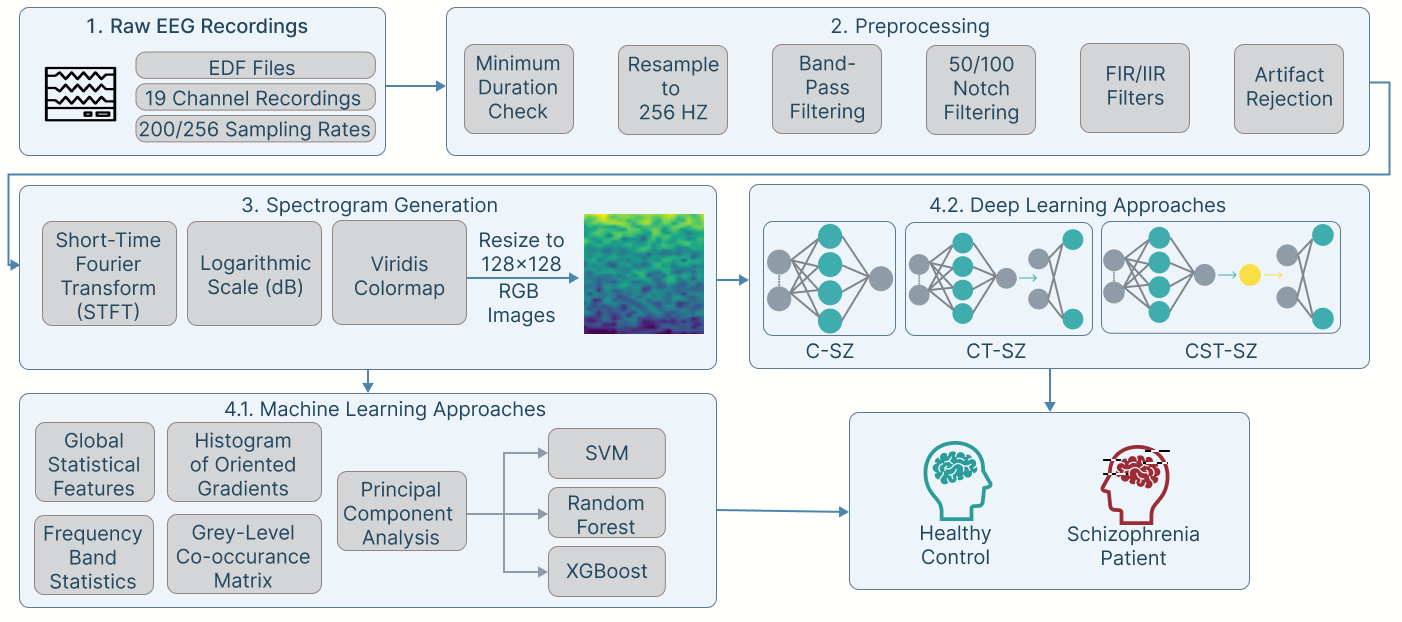}
    \caption{Proposed framework}
    \label{fig:framework}
\end{figure*}

Figure~\ref{fig:framework} shows the proposed framework. It consists of preparing dataset recordings, preprocessing the input vectors, spectrogram generation, and finally, Schizophrenia detection.

\subsection{Raw EEG Recordings' Dataset}
\label{sec:data}
This work utilizes the African Schizophrenia EEG Dataset, ASZED-153~\cite{mosaku25}, which was collected in two medical institutions in Nigeria, and contains 19-channel recordings of 153 subjects, where 76 subjects belong to the case subset and 77 subjects belong to the control subset. {Table~\ref{tab:datastat} shows a summary of dataset properties.} In comparison to other datasets on Schizophrenia disease~\cite{olejarczyk17}, ASZED-153 is a larger and richer cohort for Schizophrenia detection. 
\begin{figure}[htbp]
    \centering
    \includegraphics[width=0.4\textwidth]{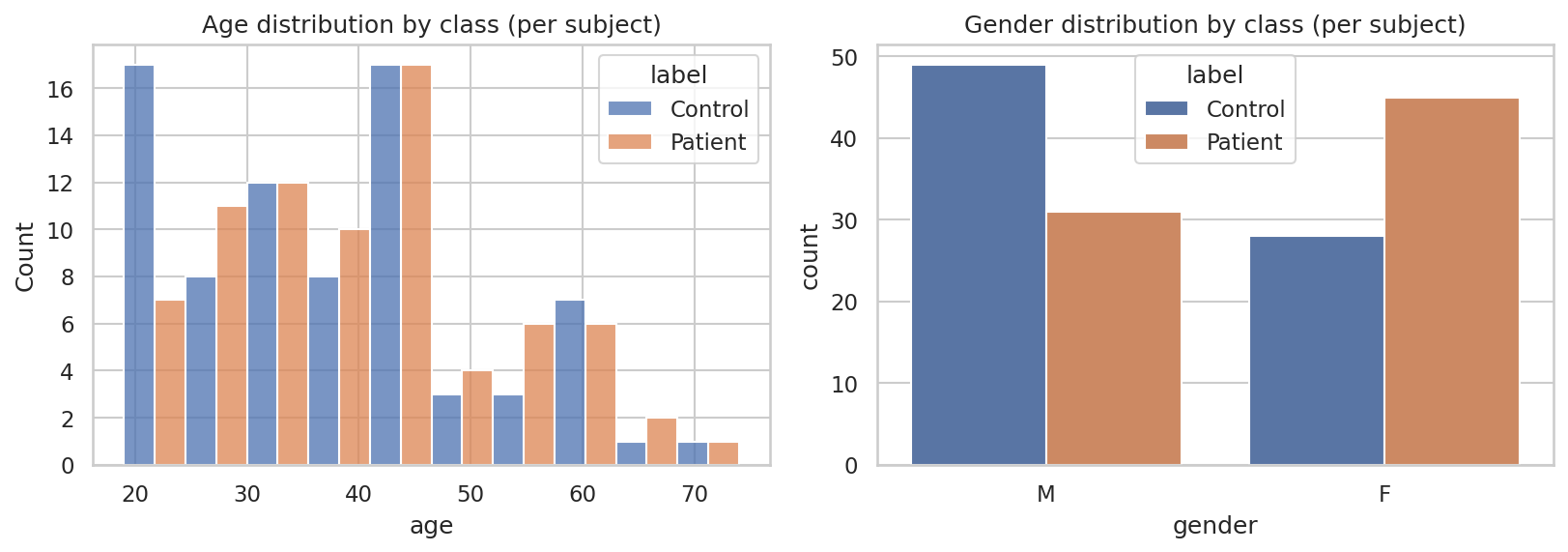}
    \caption{ASZED-153 demography, including age and gender.}
    \label{fig:demo}
\end{figure}

\begin{table}[ht]
\centering
\caption{Summary of the ASZED-153 dataset.}
\label{tab:datastat}
\begin{tabular}{ll}
\toprule
\textbf{Characteristic} & \textbf{Value} \\
\midrule
Subjects & 153 \\
Patients / Controls & 76 / 77 \\
EEG Channels & 19 \\
Sampling Rate & 200 Hz and 256 Hz \\
Recording Type & Resting-state EEG \\
Data Format & EDF \\
Classification Task & Schizophrenia vs. Healthy Control \\
Population & African cohort \\
Gender & 80M/73F \\
Age & Control: $37.6\pm12.7$, patient: $40.0\pm 12.5$ \\
\bottomrule
\end{tabular}
\end{table}

Figure~\ref{fig:demo} shows demographics of the dataset including gender and age.

\subsection{Preprocessing}

As our explorations as well as related works~\cite{bhadra24} confirm, feeding raw data into the model does not produce an efficient representation for downstream tasks; therefore, we convert the raw data into images after applying a preprocessing pipeline.
% \begin{itemize}
%     \item Recording Selection and Resampling
%     \item Band-pass, Linear-Phase and notch Filtering
%     \item Artifact Rejection
% \end{itemize}
The preprocessing pipeline was applied to every EDF recording that passed the initial integrity checks, following established EEG preprocessing practices~\cite{luck14}.

Recordings shorter than 2 seconds were discarded because they do not provide sufficient data for reliable analysis. Since the dataset was acquired using two different EEG acquisition systems with different channel naming conventions and sampling frequencies ($200$ Hz and $256$ Hz), a standardized set of $19$ EEG channels was first selected. 
All recordings sampled at $200$ Hz were then resampled to $256$ Hz using MNE resampling to ensure a consistent temporal resolution across the dataset.

Next, a \textit{band-pass filter} ($0.5–40$ Hz) was applied to remove slow baseline drift and high-frequency noise while preserving the EEG frequency bands relevant to Schizophrenia analysis~\cite{gramfort13}. A \textit{linear-phase FIR filter} was used whenever the recording duration was sufficient; otherwise, an IIR filter was employed for shorter recordings. 
% This choice was necessary because the FIR filter designed for a $0.5$ Hz high-pass cutoff requires approximately $1,691$ samples at a sampling frequency of $256$ Hz (about $6.6$ s), making it unsuitable for many short recordings. 
% Since the median recording length in the ASZED dataset is approximately 7 seconds, the majority of recordings are close to this limit, and recordings shorter than the required filter length are processed using an IIR filter, which does not impose a minimum signal-length requirement while providing comparable performance for EEG classification.

To suppress power-line interference, \textit{notch filters} were applied at 50 Hz and its first harmonic (100 Hz), corresponding to the European mains frequency. \textit{Common-mode noise} was subsequently reduced using a common average reference (CAR)\cite{mcfarland1997}, a widely adopted re-referencing technique that improves the signal-to-noise ratio by removing spatially correlated noise across channels.

Finally, an \textit{artifact rejection} procedure was performed to identify recordings with excessive signal contamination. A recording was discarded if more than $20\%$ of its samples exceeded an amplitude of $\pm200$ $\mu V$. By applying these steps, the preprocessing is completed. 
% This criterion removes recordings dominated by severe artifacts while preserving those containing only brief transient disturbances, such as electrode movement or poor contact affecting a limited portion of the recording. Compared with a more conservative threshold of $\pm150$ $\mu V$ and a $10\%$ rejection criterion, the adopted threshold substantially reduces unnecessary rejection of otherwise usable recordings while still excluding heavily contaminated signals.

\subsection{Spectrogram Generation}
Time–frequency representations were generated from the preprocessed EEG signals using the \textit{Short-Time Fourier Transform} (STFT) with a Hann window of $256$ samples and $75\%$ overlap. The resulting spectral power was averaged across the $19$ selected EEG channels and converted to a logarithmic scale (dB). To improve visualization consistency, the power values were clipped to the range of $\left[-20, 40\right]$ dB and mapped using the viridis colormap. The resulting spectrograms were resized to $128\times128$ pixel PNG images, which were subsequently used as inputs for image-based classification models. Each subject contributed between 4 and 15 spectrogram images, depending on the number of available recording sessions and protocol phases. To prevent data leakage and ensure a realistic evaluation scenario, all dataset splits were performed at the subject level rather than the image level. 
% Image-level splitting could result in samples from the same subject appearing in both training and testing sets, allowing models to exploit subject-specific characteristics such as electrode placement variations, physiological noise patterns, or recording artifacts instead of learning disease-related EEG patterns. Such leakage can lead to artificially inflated performance and poor generalization to unseen subjects.

\begin{figure}[htbp]
    \centering
    \includegraphics[width=0.2\textwidth]{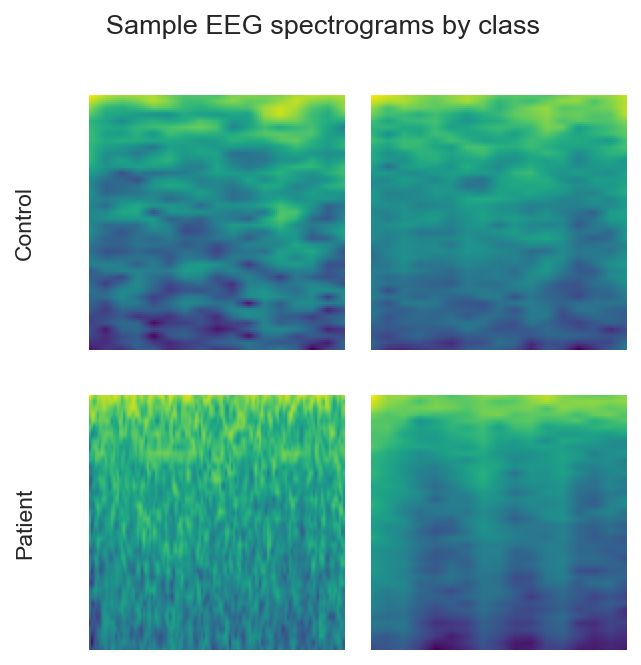}
    \caption{Generated images for case and control samples.}
    \label{fig:imgfeat}
\end{figure}
The final result of this module is the corresponding spectrogram image of the input EEG. Figure~\ref{fig:imgfeat} shows several final generated images from raw EEG signals.

\subsection{Classification Models}
% A compact CNN was developed and trained from scratch without ImageNet pretraining. 
The classification models are categorized into two categories: Machine Learning and Deep Learning approaches. For Machine Learning approaches, we utilize SVM, Random Forest (RF), and XGBoost models. 
\begin{figure*}[]
    \centering
    \includegraphics[width=0.9\textwidth]{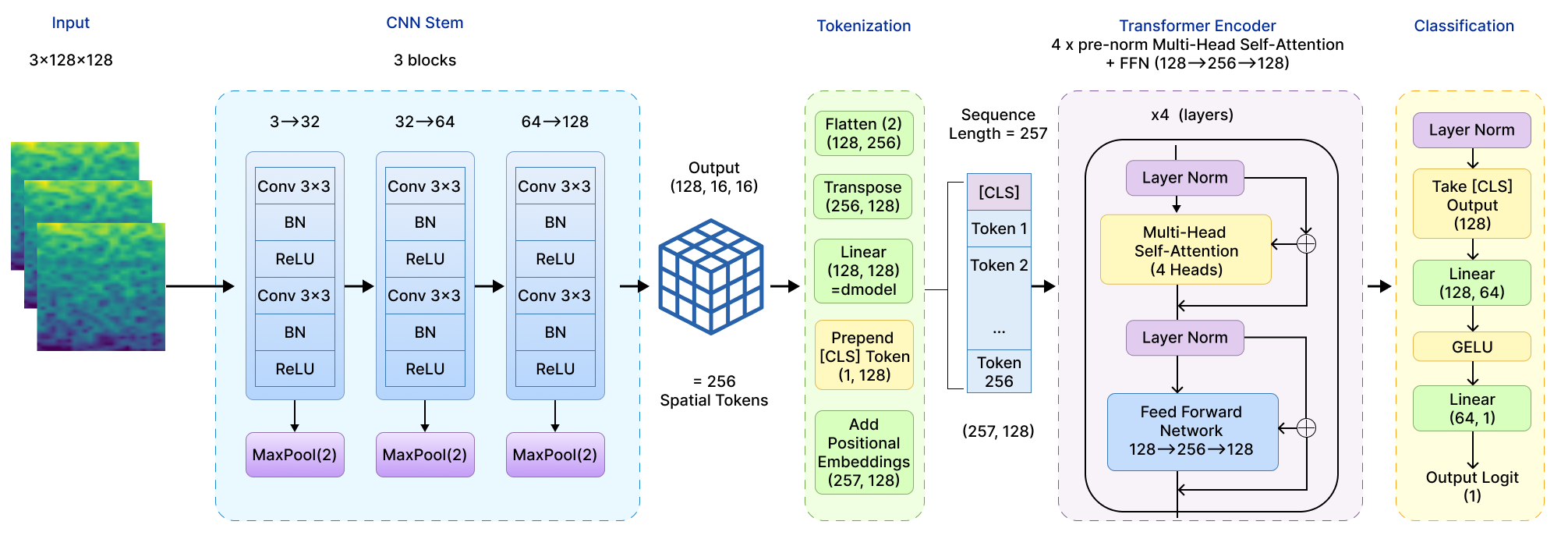}
    \caption{CT-SZ Architecture}
    \label{fig:ct-sz}
\end{figure*}
For deep models (Fig.~\ref{fig:ct-sz}), we propose three different models, which are described as follows. The first deep model is~\textit{C-SZ} which uses a CNN with four convolutional blocks, each containing two consecutive $3\times3$ convolutional layers followed by batch normalization and ReLU activation. A $2\times2$ max-pooling layer follows each block, progressively increasing the number of feature channels from 3 to 32, 64, 128, and 256 while reducing the spatial resolution from $128\times128$ to $8\times8$.

The final feature map is aggregated using global average pooling to produce a compact feature representation, which is passed through a fully connected classifier consisting of a 256-to-64 linear layer with ReLU activation, followed by dropout ($p=0.4$), and a final linear layer for binary classification. The use of two stacked $3\times3$ convolutions in each block provides an effective receptive field equivalent to a single $5\times5$ convolution while introducing additional nonlinearities and requiring fewer parameters. Furthermore, global average pooling replaces a large fully connected head, reducing the number of trainable parameters and mitigating overfitting on the relatively small training dataset.

The next deep model is CT-SZ, which is a hybrid CNN–Transformer architecture to capture both local and global patterns in EEG spectrograms. Its convolutional layers efficiently extract low-level spatial features, while the Transformer encoder models long-range dependencies by allowing every spatial token to attend to every other token from the first attention layer. This enables relationships between distant temporal segments and frequency bands to be learned directly.
The network begins with a CNN stem consisting of three convolutional blocks, each comprising two consecutive $3\times3$ convolutional layers followed by batch normalization and ReLU activation. A $2\times2$ max-pooling layer follows each block, progressively increasing the channel dimension from 3 to 32, 64, and finally 128 while reducing the spatial resolution from $128\times128$ to $16\times16$. The resulting feature map of size $128\times16\times16$ is flattened into 256 spatial tokens, which are projected to a 128-dimensional embedding space through a linear layer. A learnable classification ([CLS]) token is prepended to the token sequence, and learnable positional embeddings are added to preserve spatial information. 
% Figure~\ref{fig:ct-sz} shows the proposed CNN+Transformer framework.
The token sequence is processed by a Transformer encoder composed of four pre-normalized encoder blocks. Each block contains a multi-head self-attention module with four attention heads followed by a feed-forward network with dimensions $128\rightarrow256\rightarrow128$. The final [CLS] representation is normalized using Layer Normalization and passed through a multilayer perceptron consisting of a 128-to-64 linear layer, GELU activation, and a final linear layer producing the binary classification output.
Compared with the baseline CNN, the training procedure additionally employs a linear learning-rate warm-up during the first five epochs to improve optimization stability during the early stages of Transformer training. To enhance model interpretability, attention maps are generated from the final Transformer block by visualizing the attention weights between the [CLS] token and spatial tokens. These maps highlight the spectrogram regions that contribute most strongly to the model's predictions.
%CNN+SE+Transformer
CST-SZ is the third deep model which investigates the effect of channel-wise feature recalibration; the CNN–Transformer architecture was extended by incorporating Squeeze-and-Excitation (SE) blocks within the CNN stem. Each SE block is inserted after a convolutional block and performs global average pooling to summarize each feature channel, followed by a lightweight bottleneck network that learns channel-wise importance weights through a sigmoid gating mechanism. The resulting weights are used to rescale the corresponding feature maps before they are passed to the next layer.

While the Transformer encoder emphasizes spatial relationships across time and frequency, the SE modules adaptively emphasize the most informative feature channels. These complementary attention mechanisms enable the network to model both where discriminative information is located and which extracted features are most relevant for classification.
% To isolate the contribution of the SE modules, the CNN–SE–Transformer model retains the same Transformer configuration, embedding dimension, number of attention heads, encoder depth, dropout settings, and training schedule as the CNN–Transformer model. 
% Consequently, any observed performance differences can be attributed directly to the addition of the SE blocks rather than changes in model capacity or optimization strategy. The SE modules introduce only a small increase in the number of trainable parameters, adding approximately 5,000–10,000 parameters to the overall network.
\begin{table}[t]
\centering
\caption{Handcrafted features extracted from spectrogram images.}
\label{tab:feature_families}
\renewcommand{\arraystretch}{0.75}
\begin{tabular}{l r}
\toprule
\textbf{Feature family} & \textbf{Dimensions} \\
\midrule
Per-channel global statistics (R, G, B) & 27 \\
Grayscale global statistics & 9 \\
Frequency-band statistics (8 horizontal bands) & 16 \\
Histogram of Oriented Gradients (HOG) & 324 \\
Gray-Level Co-occurrence Matrix (GLCM) & 12 \\
PCA-reduced flattened pixels (50 PCs) & 50 \\
\midrule
\textbf{Total} & \textbf{438} \\
\bottomrule
\end{tabular}
\end{table}
% \begin{table}[t] \centering \caption{Handcrafted features extracted from spectrogram images.} \label{tab:feature_families} \begin{tabular}{l c p{0.5cm}} \toprule \textbf{Family} & \textbf{Dimensionality} \\ \toprule Per-channel global statistics (R, G, B) & 27 \\ Grayscale global statistics & 9 \\ Frequency-band statistics (8 horizontal bands) & 16 \\ Histogram of Oriented Gradients (HOG) & 324 \\ Gray-Level Co-occurrence Matrix (GLCM) & 12 \\ PCA-reduced flattened pixels (50 PCs) & 50 \\ \bottomrule \textbf{Total} & \textbf{438} \\ \bottomrule \end{tabular} \end{table}
\subsection{Feature extraction for ML methods}
The Machine Learning approaches need an extra step of feature extraction from spectrogram features, and the resulting features with their dimensionality are summarized in Table~\ref{tab:feature_families}. The feature extraction is shown in the bottom row of Figure~\ref{fig:framework}. To do this, each spectrogram image was converted into a fixed-length feature vector composed of several complementary descriptor families. Global statistical features were first extracted from the RGB channels and the grayscale image to characterize the overall distribution of color and intensity values. Frequency-band statistics were then computed from predefined horizontal regions of the spectrogram to capture how signal energy is distributed across different frequency ranges. To encode local structural information, Histogram of Oriented Gradients (HOG) descriptors were employed, providing a representation of edge orientations and texture patterns within the image. In addition, Gray-Level Co-occurrence Matrix (GLCM) features were extracted to quantify second-order texture properties, including contrast, homogeneity, energy, and correlation. Finally, PCA was applied to the flattened pixel values to obtain a compact representation that preserves the dominant modes of image variation while reducing dimensionality. 
% The dimensionality contributed by each feature family is also listed in Table~\ref{tab:feature_families}. 
% To avoid data leakage, both feature normalization and PCA were fitted exclusively on the training set. 
% Feature standardization was performed using a StandardScaler trained on the training partition, and the learned scaling parameters and PCA transformation were subsequently applied to the validation and test sets, ensuring a consistent preprocessing pipeline throughout model training and evaluation.

\section{Results}
\label{sec:res}
This section provides the results of the proposed approaches. To do this, we performed them on a computer equipped with an Intel Core i5-210H processor, an NVIDIA GeForce RTX 4050 Laptop GPU (6 GB VRAM), and 16 GB RAM. The proposed models were implemented in Python using the PyTorch Deep Learning framework with CUDA acceleration, numpy, pandas, scipy, scikit-learn, XGBoost, pillow, scikit-image, and mne.
To compare the performance of the methods, we divided the dataset into training, validation, and test subsets using a 70\%/15\%/15\% split at the subject level, stratified by class. After preprocessing, 1,276 spectrogram images from 122 subjects were retained. The training set comprised 900 images from 86 subjects (48 control, 38 patient; 525 control and 375 patient images). The validation set comprised 184 images from 18 subjects (10 control, 8 patient; 113 and 71 images, respectively), and the test set comprised 192 images from 18 subjects (10 control, 8 patient; 125 and 67 images, respectively). Subjects from each class were divided independently and then combined into the final subsets, so no subject appears in more than one subset.
% To compare the performance of the methods, we divided the dataset into training, validation, and test subsets using a $70\%/15\%/15\%$ split at the subject level. Therefore, 1,801 images were partitioned into training, validation, and test subsets at the subject level. The training set comprised 1,252 images from 107 subjects, including 649 control and 603 patient images. The validation set consisted of 280 images from 23 subjects, with 154 control and 126 patient images, while the test set contained 269 images from 23 subjects, comprising 129 control and 140 patient images. Class distribution was preserved through stratified splitting, where subjects from each class were independently divided and subsequently combined into the final subsets. 

% The resulting 1,801 images were partitioned into training, validation, and test subsets at the subject level. The training set comprised 1,252 images from 107 subjects, including 649 control and 603 patient images. The validation set consisted of 280 images from 23 subjects, with 154 control and 126 patient images, while the test set contained 269 images from 23 subjects, comprising 129 control and 140 patient images.
\color{black}
% The integrity of the splitting procedure was verified programmatically by ensuring that no subject identifier appeared in more than one subset.
Hyperparameter optimization for the SVM and XGBoost classifiers was performed using GridSearchCV and RandomizedSearchCV with 5-fold GroupKFold Cross-Validation. 
Subject identifiers were used as grouping variables to ensure that images from the same subject were never assigned to different folds, thereby preventing subject-level data leakage during model selection. 
The hyperparameter search was guided by the area under the receiver operating characteristic curve (AUC-ROC) as the optimization metric.
The CNN network was optimized using the Adam optimizer with a class-weighted binary cross-entropy loss to compensate for class imbalance. The learning rate was adaptively adjusted using the ReduceLROnPlateau scheduler based on the validation AUPRC, and early stopping with a patience of 10 epochs was employed to prevent overfitting.

\begin{table*}[t]
\centering
\caption{Performance comparison on the test set at image and subject levels.}
\label{tab:model_results_combined}
\begin{tabular}{clccccccc}
\toprule
\textbf{Level} & \textbf{Metric} 
& \textbf{DWT-CNN} 
& \textbf{SVM} 
& \textbf{RF} 
& \textbf{XGBoost} 
& \textbf{C-SZ} 
& \textbf{CT-SZ} 
& \textbf{CST-SZ} \\
\midrule
\multirow{2}{*}{Subject-level}
& AUC-ROC
& 0.8750 & 0.9125 & 0.9250 & 0.9125 & 0.8375 & \textbf{0.9500} & 0.9250 \\
& AUPRC
& 0.8015 & 0.8835 & 0.8755 & 0.8339 & 0.7469 & \textbf{0.9318} & 0.8755 \\
\midrule
\multirow{2}{*}{Image-level}
& AUC-ROC
& 0.8240 & 0.8933 & 0.9004 & 0.8958 & 0.8841 & 0.9151 & \textbf{0.9288} \\
& AUPRC
& 0.7150 & 0.7458 & 0.7181 & 0.7364 & 0.6877 & 0.8084 & \textbf{0.8209} \\
\bottomrule
\end{tabular}
\end{table*}

% \begin{table*}[t]
% \centering
% \caption{Performance comparison on the test set at image and subject levels.}
% \label{tab:model_results_combined}

% % ---------- Subject-level results ----------

% \vspace{2mm}

% \begin{tabular}{lcccccccc}
% \toprule
% \textbf{level}
% & \textbf{Metric}
% & \textbf{DWT-CNN\cite{bhadra24}}
% & \textbf{SVM}
% & \textbf{RF}
% & \textbf{XGBoost}
% & \textbf{C-SZ}
% & \textbf{CT-SZ}
% & \textbf{CST-SZ} \\
% \midrule
% Subject-level
% & AUC-ROC
% & 0.8750
% & 0.9125
% & 0.9250
% & 0.9125
% & 0.9125
% & \textbf{0.9500}
% & 0.9250 \\

% & AUPRC
% & 0.8015
% & 0.8835
% & 0.8755
% & 0.8339
% & 0.8339
% & \textbf{0.9318}
% & 0.8755 \\
% \bottomrule
% \end{tabular}

% \vspace{4mm}

% % ---------- Image-level results ----------
% \textbf{Image Level}

% \vspace{2mm}

% \begin{tabular}{lccccccc}
% \toprule
% \textbf{Metric}
% & \textbf{DWT-CNN\cite{bhadra24}}
% & \textbf{SVM}
% & \textbf{RF}
% & \textbf{XGBoost}
% & \textbf{C-SZ}
% & \textbf{CT-SZ}
% & \textbf{CST-SZ} \\
% \midrule
% AUC-ROC
% & 0.8240
% & 0.8933
% & 0.9004
% & 0.8958
% & 0.8841
% & 0.9151
% & \textbf{0.9288} \\

% AUPRC
% & 0.7150
% & 0.7458
% & 0.7181
% & 0.7364
% & 0.6877
% & 0.8084
% & \textbf{0.8209} \\
% \bottomrule
% \end{tabular}

% \end{table*}
Two types of prediction were obtained, i.e., image level and subject level. The subject-level prediction was obtained by averaging the predicted probabilities across all spectrograms belonging to the same subject. Specifically,
$ P_{\text{subject}} = \frac{1}{N}\sum_{i=1}^{N} P_i $
where \(P_i\) is the predicted probability for the \(i\)-th spectrogram and \(N\) is the number of spectrograms for that subject. The final subject-level class was assigned using a threshold of 0.5.

\begin{figure}[htbp]
    \centering    \includegraphics[width=0.4\textwidth]{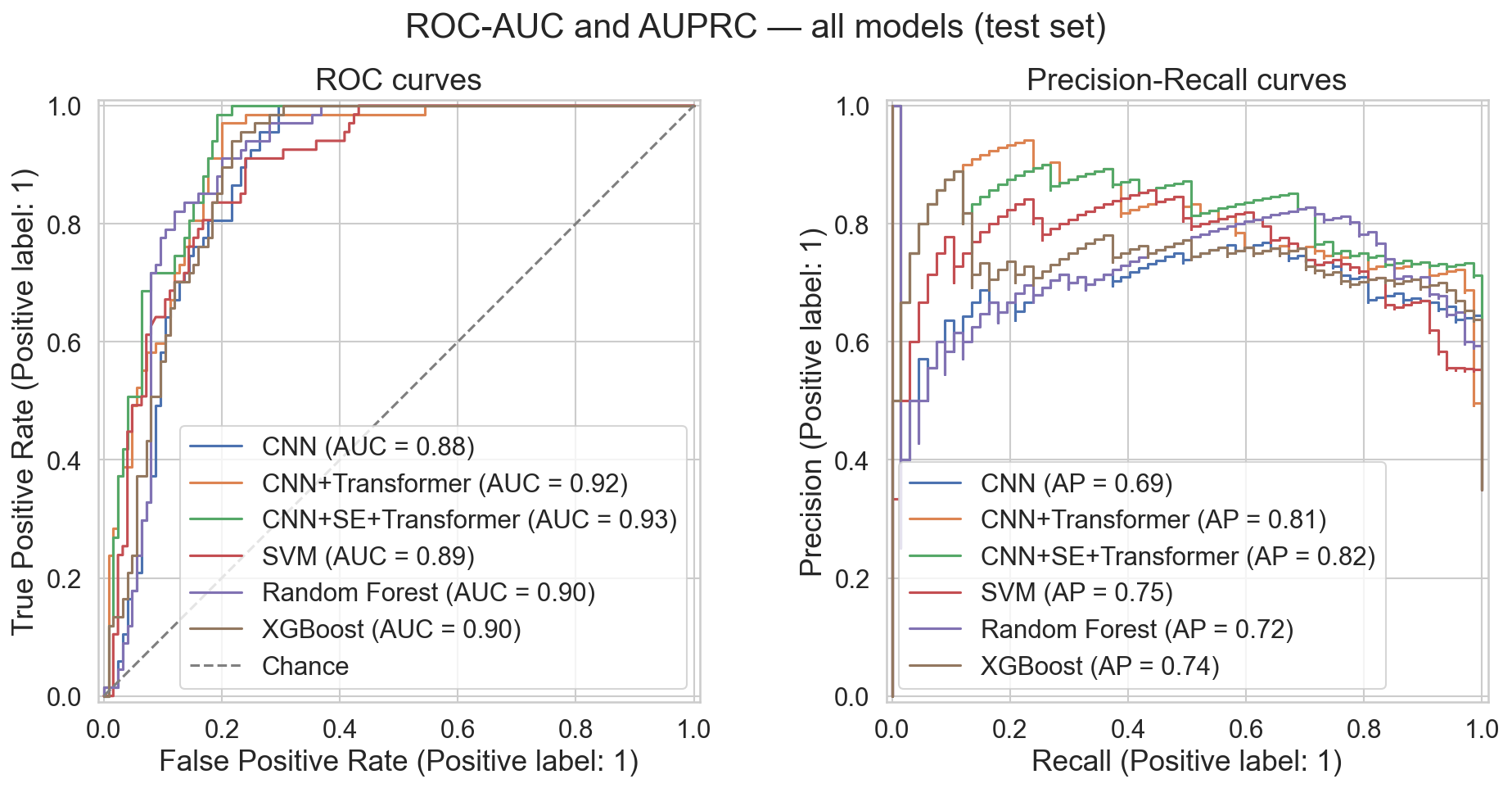}
    \caption{AUPR \& AUC-ROC Overlay.}
    \label{fig:aupr}
\end{figure}

Table~\ref{tab:model_results_combined} shows the results of the method based on the AUC-ROC and AUPRC for both image- and subject-level evaluation. 
For a fair comparison, a published DWT-based Schizophrenia detection method (DWT-CNN)~\cite{bhadra24} was reimplemented and evaluated on the ASZED-153 dataset under the same preprocessing and subject-wise evaluation protocol as the proposed models. The DWT-CNN was originally applied on IBIB PAN~\cite{olejarczyk17} with a limited number of subjects.
In addition, we implemented Machine Learning methods, i.e., SVM, RF, and XGBoost. 
As the results confirm, the proposed Transformer family with the customized data preparation of this work has the highest performance. While CST-SZ achieves the best performance in \textit{image-level}, CT-SZ has the highest scores in \textit{subject-level}. On the other hand, C-SZ, the convolution-based approach, has the lowest performance among the proposed approaches. 
Figure~\ref{fig:aupr} shows an AUPR and AUC-ROC overlay for image-level data handling.

% Although the original work demonstrated promising results for EEG-based classification, the study did not explicitly describe the use of subject-level data partitioning. Therefore, it is possible that the reported evaluation was performed using window-level or sample-level splitting, which may introduce data leakage when multiple segments from the same subject are distributed across training and testing sets. Such leakage can result in overly optimistic performance estimates by allowing subject-specific information to be indirectly shared between subsets. 

%==============================================================
% \begin{figure}[htbp]
%     \centering
%     \includegraphics[width=0.3\textwidth]{Fig/pr_overlay.png}
%     \caption{AUPR Overlay.}
%     \label{fig:aupr}
% \end{figure}
% Figure~\ref{fig:aupr} shows AUPR overlay
%==============================================================
%==============================================================
% \begin{figure}[htbp]
%     \centering
%     \includegraphics[width=0.3\textwidth]{Fig/roc_overlay.png}
%     \caption{AUC-ROC Overlay.}
%     \label{fig:roc}
% \end{figure}
% Figure~\ref{fig:roc} shows AUC-ROC overlay
%==============================================================

\section{Conclusion}
\label{sec:conc}
This study has presented an automated framework for Schizophrenia classification using EEG signals transformed into time-frequency spectrogram representations. A standardized preprocessing pipeline was employed to produce consistent image-based representations suitable for Machine Learning. The generated spectrograms were evaluated using both classical Machine Learning algorithms as well as Deep Learning models based on Convolutional Neural Networks and hybrid CNN–Transformer architectures. To ensure a reliable and clinically meaningful evaluation, all experiments were conducted using subject-level data partitioning, thereby mitigating the possibility of data leakage caused by samples from the same subject appearing in multiple subsets. The experimental results demonstrate that EEG-derived spectrograms contain discriminative information for automated Schizophrenia detection and that both traditional Machine Learning and Deep Learning approaches can effectively exploit these representations. Furthermore, the use of a standardized preprocessing pipeline and subject-level evaluation provides a more realistic assessment of model generalization than sample-level validation strategies commonly adopted in previous studies.
Future work will focus on improving classification performance through advanced time-frequency representations, self-supervised and Transformer-based architectures, multimodal EEG feature fusion, and validation on larger, multi-center EEG datasets. These directions may contribute to the development of more robust and clinically applicable computer-aided diagnostic systems for Schizophrenia.
\bibliographystyle{unsrt} % We choose the "plain" reference style
\bibliography{ref} % Entries are in the refs.bib file

@article{sravanthi26,
  title={Schizophrenia detection via lobe-wise and overall EEG features using VMD and bayesian-optimized machine learning models},
  author={Sravanthi, Gandham Sai and Sharma, Lakhan Dev},
  journal={Frontiers in Neuroscience},
  volume={20},
  pages={1753779},
  year={2026},
  publisher={Frontiers Media SA}
}

@article{jahmunah19,
  title={Automated detection of schizophrenia using nonlinear signal processing methods},
  author={Jahmunah, V and Oh, Shu Lih and Rajinikanth, V and Ciaccio, Edward J and Cheong, Kang Hao and Arunkumar, N and Acharya, U Rajendra},
  journal={Artificial intelligence in medicine},
  volume={100},
  pages={101698},
  year={2019},
  publisher={Elsevier}
}

@article{zhang26,
  title={Global lifetime prevalence of schizophrenia: A systematic review and meta-analysis},
  author={Zhang, Shuxin and Chen, Yubin and Zhang, Linghui and Yang, Xinyu and Dong, Junhui and Lu, Minle and Zhou, Na and Feng, Yang and Zhang, Yu and Zhou, Yuqiu},
  journal={Molecular Psychiatry},
  pages={1--12},
  year={2026},
  publisher={Nature Publishing Group UK London}
}

@article{rahul24,
  title={A systematic review of EEG based automated schizophrenia classification through machine learning and deep learning},
  author={Rahul, Jagdeep and Sharma, Diksha and Sharma, Lakhan Dev and Nanda, Umakanta and Sarkar, Achintya Kumar},
  journal={Frontiers in Human Neuroscience},
  volume={18},
  pages={1347082},
  year={2024},
  publisher={Frontiers Media SA}
}

@book{luck14,
  title={An introduction to the event-related potential technique},
  author={Luck, Steven J},
  year={2014},
  publisher={MIT press}
}

@article{gramfort13,
  title={MEG and EEG data analysis with MNE-Python},
  author={Gramfort, Alexandre and Luessi, Martin and Larson, Eric and Engemann, Denis A and Strohmeier, Daniel and Brodbeck, Christian and Goj, Roman and Jas, Mainak and Brooks, Teon and Parkkonen, Lauri and others},
  journal={Frontiers in Neuroinformatics},
  volume={7},
  pages={267},
  year={2013},
  publisher={Frontiers Media SA}
}

@article{mcfarland1997,
  title={Spatial filter selection for EEG-based communication},
  author={McFarland, Dennis J and McCane, Lynn M and David, Stephen V and Wolpaw, Jonathan R},
  journal={Electroencephalography and clinical Neurophysiology},
  volume={103},
  number={3},
  pages={386--394},
  year={1997},
  publisher={Elsevier}
}

@inproceedings{woo18,
  title={Cbam: Convolutional block attention module},
  author={Woo, Sanghyun and Park, Jongchan and Lee, Joon-Young and Kweon, In So},
  booktitle={Proceedings of the European conference on computer vision (ECCV)},
  pages={3--19},
  year={2018}
}

@article{aich25,
  title={Schizophrenia detection from electroencephalogram signals using image encoding and wrapper-based deep feature selection approach},
  author={Aich, Utathya and Saha, Arghyasree and Wo{\'z}niak, Marcin and Ijaz, Muhammad Fazal and Singh, Pawan Kumar},
  journal={Scientific Reports},
  volume={15},
  number={1},
  pages={21390},
  year={2025},
  publisher={Nature Publishing Group UK London}
}

@article{khare21,
  title={SPWVD-CNN for automated detection of schizophrenia patients using EEG signals},
  author={Khare, Smith K and Bajaj, Varun and Acharya, U Rajendra},
  journal={IEEE Transactions on Instrumentation and Measurement},
  volume={70},
  pages={1--9},
  year={2021},
  publisher={IEEE}
}

@article{hossain26,
  title={Exploring brain lobe-specific insights in an explainable framework for EEG-based schizophrenia detection},
  author={Hossain, Md Milon and Tawhid, Md Nurul Ahad},
  journal={PLoS One},
  volume={21},
  number={3},
  pages={e0334389},
  year={2026},
  publisher={Public Library of Science San Francisco, CA USA}
}

@article{jangde26,
  title={EEG-Based Schizophrenia Classification Using Attention-Integrated Deep Convolutional Networks},
  author={Jangde, Anjali Sagar and Verma, Gyanendra Kumar},
  journal={Psychiatry Research: Neuroimaging},
  pages={112138},
  year={2026},
  publisher={Elsevier}
}

@inproceedings{bhadra24,
  title={Automatic Schizophrenia Detection Using Discrete Wavelet Transform from EEG Signal},
  author={Bhadra, Sweta and Kumar, Chandan Jyoti},
  booktitle={NIELIT's International Conference on Communication, Electronics and Digital Technologies},
  pages={541--558},
  year={2024},
  organization={Springer}
}

@article{olejarczyk17,
  title={Graph-based analysis of brain connectivity in schizophrenia},
  author={Olejarczyk, Elzbieta and Jernajczyk, Wojciech},
  journal={PloS one},
  volume={12},
  number={11},
  pages={e0188629},
  year={2017},
  publisher={Public Library of Science San Francisco, CA USA}
}

@article{mosaku25,
  title={An open-access EEG dataset from indigenous African populations for schizophrenia research},
  author={Mosaku, SK and Olateju, EO and Ayodele, KP and Akinsulore, A and Ajiboye, PO and Oloniniyi, OI and Ayorinde, A and Agboola, O and Obayiuwana, E and Akinwale, OB and others},
  journal={Data in Brief},
  pages={111934},
  year={2025},
  publisher={Elsevier}
}
% \bibitem{b1} G. Eason, B. Noble, and I. N. Sneddon, ``On certain integrals of Lipschitz-Hankel type involving products of Bessel functions,'' Phil. Trans. Roy. Soc. London, vol. A247, pp. 529--551, April 1955.
% \bibitem{b2} J. Clerk Maxwell, A Treatise on Electricity and Magnetism, 3rd ed., vol. 2. Oxford: Clarendon, 1892, pp.68--73.
% \bibitem{b3} I. S. Jacobs and C. P. Bean, ``Fine particles, thin films and exchange anisotropy,'' in Magnetism, vol. III, G. T. Rado and H. Suhl, Eds. New York: Academic, 1963, pp. 271--350.
% \bibitem{b4} K. Elissa, ``Title of paper if known,'' unpublished.
% \bibitem{b5} R. Nicole, ``Title of paper with only first word capitalized,'' J. Name Stand. Abbrev., in press.
% \bibitem{b6} Y. Yorozu, M. Hirano, K. Oka, and Y. Tagawa, ``Electron spectroscopy studies on magneto-optical media and plastic substrate interface,'' IEEE Transl. J. Magn. Japan, vol. 2, pp. 740--741, August 1987 [Digests 9th Annual Conf. Magnetics Japan, p. 301, 1982].
% \bibitem{b7} M. Young, The Technical Writer's Handbook. Mill Valley, CA: University Science, 1989.
% \end{thebibliography}

\end{document}